\documentclass[sn-mathphys]{sn-jnl}

\jyear{2023}%

\theoremstyle{thmstyleone}%
\theoremstyle{thmstyletwo}%

\theoremstyle{thmstylethree}%

\begin{document}

\title[A Multilingual NL2SQL with Database Schema Pruning to Improve Self-attention]{
A Multilingual Translator to SQL with Database Schema Pruning to Improve Self-Attention}


\author*[1,2]{\fnm{Marcelo} \sur{Archanjo Jose}}\email{marcelo\_archanjo@yahoo.com.br}\equalcont{These authors contributed equally to this work.}

\author[2,3]{\fnm{Fabio} \sur{Gagliardi Cozman}}\email{fgcozman@usp.br}
\equalcont{These authors contributed equally to this work.}

\affil*[1]{\orgdiv{Institute of Advanced Studies}, \orgname{University of São Paulo}, \orgaddress{\street{R. do Anfiteatro, 513}, \city{São Paulo}, \postcode{05508-060}, \state{São Paulo}, \country{Brazil}}}

\affil[2]{\orgdiv{Center for Artificial Intelligence}, \orgname{C4AI}, \orgaddress{\street{Av. Prof. Lúcio Martins Rodrigues, 370}, \city{São Paulo}, \postcode{05508-010}, \state{São Paulo}, \country{Brazil}}}

\affil[3]{\orgname{Escola Polit\'ecnica, University of São Paulo}, \orgaddress{\street{Av. Professor Luciano Gualberto 2231}, \city{São Paulo}, \postcode{05508-010}, \state{São Paulo}, \country{Brazil}}}


\abstract{Long sequences of text are challenging in the context of transformers,
due to quadratic memory increase in the self-attention mechanism. 
As this issue directly affects the translation from natural language to SQL queries
(as techniques usually take as input a concatenated text with the question and the database schema),
we present techniques that allow long text sequences 
to be handled by transformers with up to 512 input tokens. 
We propose a training process with database schema pruning (removal of
tables and columns names that are useless for the query of interest). 
In addition, we used a multilingual approach with the mT5-large model fine-tuned with a data augmented Spider dataset in four languages simultaneously: English, Portuguese, Spanish and French. Our 
proposed technique used the Spider dataset and increased the exact set match accuracy results from 0.718 to 0.736 in a validation dataset (Dev). Source code, evaluations, and checkpoints are available at: \underline{https://github.com/C4AI/gap-text2sql}.}




\maketitle

\section{Introduction}\label{sec1}

Transformers  with the attention mechanism have led to great leaps in natural language processing (NLP)~\cite{Attention2017}. However, they do have limitations. An example is the 512 tokens input limit,
as this can be a drawback when dealing with long text sequences. The number of tokens is not really a limitation as it can be increased; however, expanding this number  increases memory consumption quadratically and may disperse attention through many tokens. 
Different proposals, such as Big Bird~\cite{BigBird2020}, Longformer~\cite{Longformer2020}, Poolingformer~\cite{Poolingformer2021}, ETC~\cite{ETC2020}, Linformer~\cite{Linformer2020}, Reformer~\cite{Reformer 2020}, among others,  process long text sequences and address the challenge of memory consumption by letting it grow near linearly, while keeping good performance.


In this paper we explore techniques that enhance transformers in the context
of {\em natural language to SQL} (NL2SQL) translation.
Existing NL2SQL parsers encode the combined text composed of the question and database schema information, especially the table names, column names, and their relations. 
More information about NL2SQL can be found in these surveys: ~\cite{SurveyYu2010}~\cite{SurveyKim2020}~\cite{SurveyOzcan2020}.

NL2SQL parsers based on transformers have greatly evolved in the last few years. The Spider dataset\footnote{Spider 
dataset: \underline{https://yale-lily.github.io/spider}.}~\cite{Yu2018b} has had a key role in that progress
due to its features, such as the number of databases, query complexity, etc. 
The current leaderboard (measuring exact set match without values) is presented in Table~\ref{tabLeaderboard}.

\begin{table}[t]
\begin{center}
\begin{minipage}{270pt}
\caption{Spider Leaderboard - Exact Set Match without Values in September 2022}\label{tabLeaderboard}%
\begin{tabular}{@{}llllc@{}}
\toprule
Rank  & Model & Test & Dev & Reference\\
\midrule
1 & Graphix-3B + PICARD & 0.740 & 0.771 & Anonymous\\
1 & CatSQL + GraPPa & 0.739 & 0.786 & Anonymous\\
3 & SHiP + PICARD & 0.731 & 0.772 & Anonymous\\
4 & G³R + LGESQL + ELECTRA & 0.726 & 0.772 & Anonymous\\
6 & RESDSQL+T5-1.1-lm100k-xl & 0.724 & 0.781 & Anonymous\\
6 & T5-SR & 0.724 & 0.772 & Anonymous\\
7 & S²SQL + ELECTRA & 0.721 & 0.764 & \cite{SSSQL2022}\\
8 & LGESQL + ELECTRA & 0.720 & 0.751 & \cite{LGESQL2021}\\
9 & T5-3B+PICARD & 0.719 & 0.755 & \cite{PICARD2021}\\

\botrule
\end{tabular}
\item * Techniques that currently do not have a paper associated are presented as anonymous.
\end{minipage}
\end{center}
\end{table}

Currently, the works in the first positions that have a paper explaining their approach are: 

- S²SQL~\cite{SSSQL2022} is a technique that injects syntactic information of the question in the encoder, rather than just the question text. They also introduce a decoupling constraint in order to induce diverse edge embedding learning.

- The idea behind LGESQL (Line Graph Enhanced Text-to-SQL)~\cite{LGESQL2021} is to use line graphs to include local (1-hop relation) and non-local (extracted from parameter matrix) features to compute. The line graph relates question nodes, table nodes, and column nodes. A graph pruning process helps indicate the relevant graph schema to the related question. The pretrained language model (PLM) ELECTRA-large-discriminator has achieved the best result.

- The PICARD~\cite{PICARD2021} (Parsing Incrementally for Constrained Auto-Regressive Decoding) approach  constrains the decoded tokens during the inference process to find valid SQL queries through four levels in the parsing process. The best result within this approach has been  achieved with the T5-3b model.

A technique that has been a reference for many other techniques with good results in the Spider leaderboard is the RAT-SQL (Relation-Aware Transformer SQL)~\cite{RAT-SQL2019} that explored a database schema link with the natural language questions words, with important results when launched.
RAT-SQL+GAP scheme~\cite{RAT-SQL+GAP2020} (0.697 Test and 0.718 Dev) is a variant that is used in this paper as a baseline.
GAP means Generation-Augmented Pre-Training; it employs a custom pre-training in the BART model with learning objectives related to the NL2SQL task. Such training increases performance when this model is plugged into the RAT-SQL parser.  

When using transformers, the limitation over long input text sequences strikes. 
The natural language question is not a problem as it is usually short; 
however, the database schema may be large depending on the number of tables and columns.

We  here use the RAT-SQL+GAP when the model is BART-large (the pretrained model version was downloaded from Github\footnote{RAT-SQL+GAP github:\underline{https://github.com/awslabs/gap-text2sql}.} ) and our multilingual mRAT-SQL version, but without GAP, when the model is mT5-large, which means the model is the original form of Hugging Face\footnote{Google's mT5:\underline{https://huggingface.co/google/mt5-large}.}.

The proposal for this paper is to present to the scientific community the improvement obtained with schema pruning in a multilingual approach. The motivation was the benefits of schema pruning because in NL2SQL with transformers it is an open problem to handle databases with big schemas that produce long input sequences that exceed 512 tokens. The multilingual approach producing better results was a good side effect, that we notice due to our previous work~\cite{mRATSQLGAP2020} using a combination of English and Portuguese languages, which was expanded with Spanish and French languages here. It is important to report and present the results of these finds to allow other researchers to analyze them as a possible choice to be applied in their context.

The main contributions of this paper are the schema pruning to reduce the number of tokens to fit in the 512, preserving the relevant tables and column names used in the queries for the corresponding database, and a multilingual data augmentation process with four languages: English, Portuguese, Spanish, and French. Both contributions can be easily incorporated into other NL2SQL approaches thus making them a viable path to increase benchmark results.

\section{Multilingual Data Augmentation}
Natural language processing now a days as great advances, but mainly in English Language. Operate with different languages could be problematic due to the limitation of language models pretrained in those languages. Multilingual language models are a good option ~\cite{mRATSQLGAP2020}~\cite{MultiSpider2021} ~\cite{MUCE2022}.

In previous work on NL2SQL in Portuguese ~\cite{mRATSQLGAP2020}, we have found that it is better to work with multilingual models than with a model for a specific language (mostly because SQL queries naturally contain many English words).
This was shown in our previous work~\cite{mRATSQLGAP2020} with the multilingual model mBART-50. Multilingual models allow training in English and Portuguese separately,  and also the two languages together. We produced better results when training the model with multiple languages than with a single one even working with the English language. It is possible to deduce this is an effect of data augmentation because we double the dataset.

In the current work, we chose mT5~\cite{mT52021} multilingual because achieves better results than mBART-50. Specifically, the mT5 large multilingual model with 1.2 billion parameters pre-trained with 101 languages, including the languages we are currently working on English, Portuguese, Spanish, and French.

The Spider dataset consists of 3 files: train\_spider.json (7,000 questions), train\_others (1,659 questions) (both train dataset), and dev.json (1,034 questions) (validation dataset).

We translated natural language questions from the Spider Dataset into Portuguese, Spanish and French and created versions of the four languages, with the same corresponding original query. We choose not to translate any information about the database schema to make the results compatible and comparable, which means we can make inferences with any of the four languages, and the resultant query can be evaluated with the Spider test suite~\cite{SpiderTestSuite2021}\footnote{Spider 
test suite: \underline{https://github.com/taoyds/test-suite-sql-eval}.}. In Table~\ref{tabQuestion} the question ``What are the maximum and minimum budgets of the departments?" are presented in four languages; all are related to the same query: ``SELECT max(budget\_in\_billions),  min(budget\_in\_billions) FROM department".
The translations were made using the  Google Translate service. 

We also created a dataset version that joins the four languages together. The original Spider has 8659 train and 1034 validation examples. This quad dataset has 34636 train and 4136 validation examples. Ours is a data augmentation approach that works with multilingual models.

\begin{table}[t]
\begin{center}
\begin{minipage}{\textwidth}
\caption{Question sample in English, Portuguese, Spanish and French, related to the same query: ``SELECT max(budget\_in\_billions) ,  min(budget\_in\_billions) FROM department"}\label{tabQuestion}%
\begin{tabular}{@{}ll@{}}
\toprule
Language  & Question\\
\midrule
English & What are the maximum and minimum budgets of the departments?\\
Portuguese & Quais são os orçamentos máximo e mínimo dos departamentos?\\
Spanish & ¿Cuáles son los valores máximo y mínimo presupuesto de los departamentos?\\
French & Quels sont le budget maximum et minimum des départements?\\
\botrule
\end{tabular}

\end{minipage}
\end{center}
\end{table}



\section{Schema pruning}

The transformer self-attention mechanism size limitation also applies to NL2SQL.

\begin{figure}[t]
\centering\includegraphics[scale=0.30]{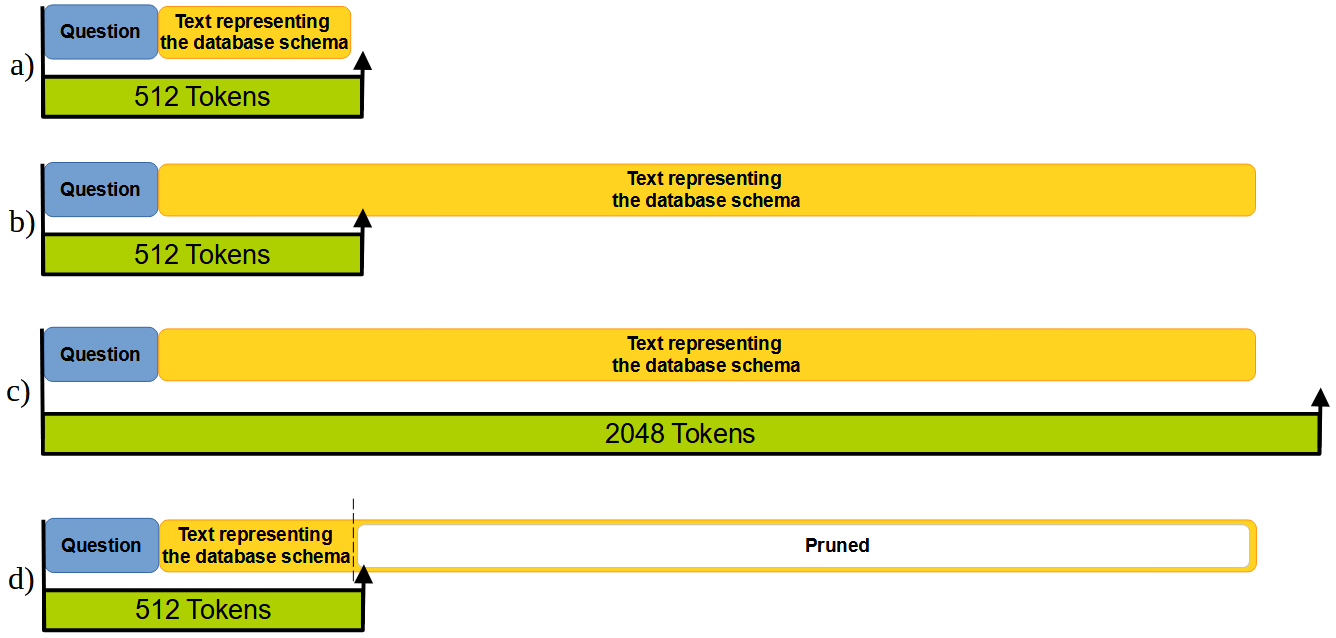}
\caption{Situations and possibles solutions; a) Ideal situation; b) Example that overcomes the 512 tokens; c) Possible solution, expand the token limit to 2048 tokens; d) Proposed solution, prune the database schema to fit in 512 token limit.} \label{fig_limits}
\end{figure}

The figure~\ref{fig_limits} presents a graphical representation of the problem. The figure~\ref{fig_limits}a shows the ideal situation where the junction of the question text and the serialized text of the schema (tables names, columns names, and their relations) fits under the 512 tokens. The figure~\ref{fig_limits}b shows a real example case where the text that represents the database schema overcomes the limit of 512 tokens. One possible solution for that situation was to expand the limit to 2048 tokens to fit all necessary text, the  figure~\ref{fig_limits}c illustrates this solution.

The problem is not the natural language part (the question) but the database schema. It is not usual for a question to have too many words (more than 512 tokens), but databases can have many tables and columns that lead to a schema with more than 512 tokens (when serialized as text).
However, one question will typically not require information of the entire database schema to generate the expected SQL query. Considering the training data set, even a group of questions may not require the entire database schema. It is possible to analyze all questions in the training dataset related to the same database and see which tables and columns are used.

This idea allows pruning table and column names that are not used for that pack of questions, reducing in that way the size of the database schema.  With this reduced version of this database schema, It is possible to fit the natural language question and the database schema under 512 tokens, respecting the self-attention mechanism limitation. The figure~\ref{fig_limits}d presents the effect of the schema pruning.

Currently, the Spider dataset is composed of 166 databases 146 for training and 20 for validation. The schemas are organized in the tables.json file. RAT-SQL-GAP has a pre-process step that prepares the information for training and inference steps. 

We noticed that the number of training questions in fact used during the RAT-SQL-GAP pre-processing are always smaller than the number of questions in the training dataset. This is due to RAT-SQL+GAP code dropping examples that have more than 512 tokens.

The original English training dataset has 8659 examples, but just 8558 were really used, using BartTokenizer, which will interfere with the number of tokens. 101 examples were rejected due to the combination of the question and the database schema (table names and column names) being greater than 512 tokens. The quad (English, Portuguese, Spanish, and French together) training dataset has 34363 examples, but just 33927 were really used, using MT5Tokenizer. 709 examples were rejected. The number of rejected examples did not increase by four times, although the quad dataset was four times larger, because different languages produce questions with a different number of words and use a different tokenizer (MT5Tokenizer).

We analyzed the rejected examples and organized them by the database in Table~\ref{tabDrops}.

\begin{table}[t]
{
\centering
\begin{center}
\begin{minipage}{250pt}
\caption{Number of examples dropped from the training dataset during pre-processing.}\label{tabDrops}%
\begin{tabular}{@{}lll@{}}
\toprule
Database &  Original dataset & Quad Dataset\\
 &  BartTokenizer & MT5Tokenizer\\
\midrule
Baseball\_1 &  82 & 328\\
cre\_Drama\_Workshop\_Groups &  19 & 328\\
Soccer\_1 & 0 & 53\\
\botrule
Total rejected & 101 & 709\\
\end{tabular}

\end{minipage}
\end{center}
}
\end{table}



The questions related to these three databases are all in the training dataset file train\_spider.json.

In order to understand the reasons why just these three databases are related to the problem of exceeding the limit of 512 tokens, the number of original tables and columns was analyzed, see Table~\ref{tabDrops}. These databases have a large number of tables and columns, for example, the Baseball\_1 DB has 353 columns; if each column name uses two unique words, the 512 tokens limit is exceeded merely by the names of the columns.

We created a code to analyze all queries related to these databases and present the tables that are really used, allowing the deletion of tables not used. To validate the process of pruning, we did it manually using the DB Browser for SQLite. First, we deleted tables not used in the queries indicated by the code; later the not-used columns when the deletion of tables was not enough. For column deletion, we did not develop a specific code, the deletion was made based on the name of the column and on a visual inspection of the queries that used the table related.

After the pruning, we updated the tables.json file with a new section that reflects the modified database. The dataset file train\_spider.json has indexes related to the original tables and columns. We create a code to update these indexes with the modified pruned version. Table~\ref{tabSizes} shows the new numbers of tables and columns in the pruned version.

This approach was made to evaluate the effects of using the entire Spider dataset (without drop examples); it can be applied to other training datasets, despite the manual effort, because it is one-time work. Creating an algorithm that makes the schema pruning automatically for each drop example candidate (more than 512 tokens) is an option, but it is worth considering that the schema pruning for each example drop candidate will create a new schema per example, not per database. To make the pruning schema per database, it is necessary to aggregate all the drop example candidates (more than 512 tokens) and relate the database candidates to make the pruning considering all the queries (with the tables and columns used).

\begin{table}[t]
\centering
\begin{center}
\begin{minipage}{320pt}
\caption{Tables and columns sizes for rejected databases before and after schema pruning.}\label{tabSizes}%
\begin{tabular}{@{}lllll@{}}
\toprule
Database & Tables & Columns & Tables & Columns\\
 & Original & Original & after pruning & after pruning\\
\midrule
baseball\_1 &  26 & 353 & 13 & 87\\
cre\_Drama\_Workshop\_Groups &  18 & 100 & 15 & 80\\
soccer\_1 & 7 & 87 & 5 & 57\\
\botrule
\end{tabular}

\end{minipage}
\end{center}
\end{table}


\section{Experiments and Analysis}
\subsection{Multilingual Data Augmentation}

The experiments were performed on the following equipment:
AMD Ryzen 9 3950X 16-Core Processor, 64GB RAM, 2 GPUs NVidia GeForce RTX 3090 24GB running Ubuntu 20.04.2 LTS.

\begin{figure}[t]
\centering\includegraphics[scale=0.60]{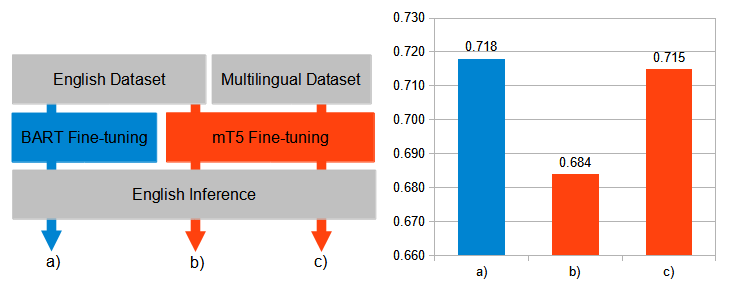}
\caption{Exact Set Match without Values, the diagram on the left, and the results on the right; a) BART-large trained in English, infer in English(baseline); b) mT5-large model trained in English, infer in English; c) mT5-large model trained in English, Portuguese, Spanish and French, infer in English.} \label{fig1}
\end{figure}

First, we reproduced the results of RAT-SQL+GAP~\cite{RAT-SQL+GAP2020} in our environment to use it as a baseline (fine tuning with 41000 steps and Batch Size=12); since it is BART-large, the GAP is active (pre-trained model by the RAT-SQL+GAP group), the train and validation datasets are in English. Figure~\ref{fig1}a shows the "Exact Set Match without Values" accuracy result 0.718, which is the same as the RAT-SQL+GAP~\cite{RAT-SQL+GAP2020} paper. This metric considers the inferred query, but not the values.
To validate the multilingual approach, we fine-tuned the mT5 model just with the original English Spider train dataset, and later with the quad (English, Portuguese, Spanish and French) Spider train dataset. Both with 51,000 steps, Batch Size=4 (this value was chosen to fit the model in our GPU memory). The validation dataset is in English for the three cases. A diagram with the three combinations is presented in Figure~\ref{fig1} on the left side. Figure~\ref{fig1}b shows an inference result of 0.864 for the model trained just in English and Figure~\ref{fig1}c shows a result of 0.715 for the model trained with the quad dataset. The three tests presented in Figure~\ref{fig1} were made with the standard self-attention of 512 tokens.

It is possible to conclude that the multilingual model mT5 produces better results when trained with more languages. The results in Figure~\ref{fig1}b and c show the increase from 0.684 to 0.715 for the same mT5-large  model first trained in English and after with the quad train dataset (English, Portuguese, Spanish and French). This increase can be credited to a data augmentation effect that was enough to make mRAT-SQL (without GAP) achieve the value of 0.715, near to the BART-large baseline of 0.718 with RAT-SQL+GAP. This makes the training process simpler because a pre-training with the model is not necessary before the final NL2SQL training.

\subsection{Schema pruning}
\begin{figure}[t]
\centering\includegraphics[scale=0.45]{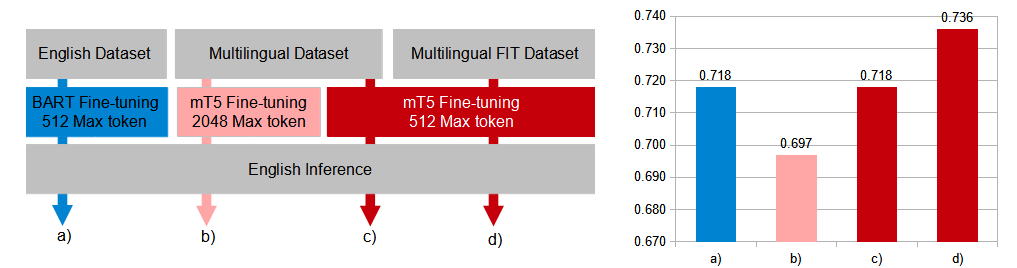}
\caption{Exact Set Match without Values, the diagram on the left, and the results on the right; a) BART-large trained in English, inferred in English (baseline); b) mT5-large model trained with the max number of token 2048 in English, Portuguese, Spanish and French, inferred in the English dataset standard; c) mT5-large model trained in English, Portuguese, Spanish and French dataset standard, inferred in English; d) mT5-large model trained in English, Portuguese, Spanish and French \textbf{Dataset FIT}, inferred in English.} \label{fig2}
\end{figure}

To understand the influence of the schema pruning we fine-tune the mT5 model with the same quad dataset (without pruning), hereinafter called "standard quad" (English, Portuguese, Spanish, and French) Spider train dataset  and after with the quad dataset with schema pruning, hereinafter called "FIT quad" (English, Portuguese, Spanish, and French) Spider train dataset. Both with 120,000 steps, Batch Size=4 and the standard self-attention of 512 tokens. We increased the number of steps to analyze as the model will converge with more steps than the 51,000 used in the prior tests, mainly because the training with mt5-large and the quad dataset achieved the best checkpoint in the last step (on an average rising slope). The validation dataset is in English for both cases. Figure~\ref{fig2}c shows the inference result of 0.718 for the model trained with the standard quad dataset and Figure~\ref{fig2}d shows results of 0.736 for the model trained with FIT quad database, whereby the schema was pruned.

The increase in fine-tuning steps indicates to be adequate; the best checkpoints were 77,500 for the standard quad dataset and 105,100 for the FIT quad train dataset. 

Another possible approach to include all text sequences during the fine-tuning process is to increase the max number of tokens in the transformer self-attention mechanism. In our case, for using the whole standard quad train dataset, it was necessary to increase from 512 to 2048 tokens. Due to the memory consumption, we had to reduce the Batch Size to just 1, but it was necessary to increase the number of steps to 480000 to get a good convergence in the model training. Figure~\ref{fig2}b shows the inference result of 0.697.

The use of the FIT quad Spider train dataset had a huge influence on the results raising from 0.718 Figure~\ref{fig2}c to 0.736 Figure~\ref{fig2}d. It can be deduced that the integral use of the training dataset, without the exclusions caused by exceeding 512 tokens, provided the best training samples. The attempt to increase the limit of 512 tokens to 2048 does not produce good results of 0.697 Figure~\ref{fig2}b. In fact, it was worst than 0.718. Figure~\ref{fig2}c  achieved by the mt5-large fine-tuned with the standard quad train dataset. The possible cause is that the attention mechanism became too sparse. A diagram with the four combinations is presented in Figure~\ref{fig2} on the left side.

\begin{table}[t]
\begin{center}
\begin{minipage}{\textwidth}
\caption{Difficult levels for the exact set match without values.}\label{tabLevels}%
\begin{tabular}{@{}lllllllll@{}}
\toprule
 & Max &   &  &   &  &   &  & \\
Model  & Token & Train  & Eval & Easy  & Medium  & Hard  & Extra  & All\\
 & Limit &   &  &   &  &   &  & \\
\midrule
Bart-large & 512 & en   & en & 0.899 & 0.744 & 0.667 & 0.428 & 0.718\\
mT5-large & 2048 & en-pt-es-fr   & en & 0.855 & 0.753 & 0.540 & 0.476 & 0.697\\
mT5-large & 512 & en-pt-es-fr   & en & 0.879 & 0.756 & 0.580 & 0.518 & 0.718\\
mT5-large & 512 & FIT-en-pt-es-fr   & en & 0.895 & 0.776 & 0.603 & 0.530 & 0.736\\
\botrule
\end{tabular}
\end{minipage}
\end{center}
\end{table}



Table~\ref{tabLevels} shows the question/query example difficulty level (easy, medium, hard, and extra hard) for the exact set match without values for the four cases. The improvement of the mT5 large fine-tuned with the FIT quad train dataset can be noticed in all levels if compared with mT5 large fine-tuned with the standard quad train dataset. The specific value of the mT5 large fine-tuned with the FIT quad train dataset for extra hard examples: 0.530 is the best of all the fine-tuning we performed, yet not reported here.

The schema pruning that produced the FIT datasets shows important results, but it was just used in the training dataset because the validation dataset does not have examples requiring more than 512 tokens. It is possible to apply the same schema pruning approach to the validation dataset because we have the query related to the question to select unused tables and columns. In future real cases of NL2SQL where only the question and the database schema are available, it will be difficult to perform the schema pruning in the inference time. One option is to analyze the need of the complete schema in an inference endpoint and create a short schema version compatible with the limit of 512 tokens to get good inferences.

\subsection{Multilingual inference}
\begin{figure}[t]
\centering\includegraphics[scale=0.5]{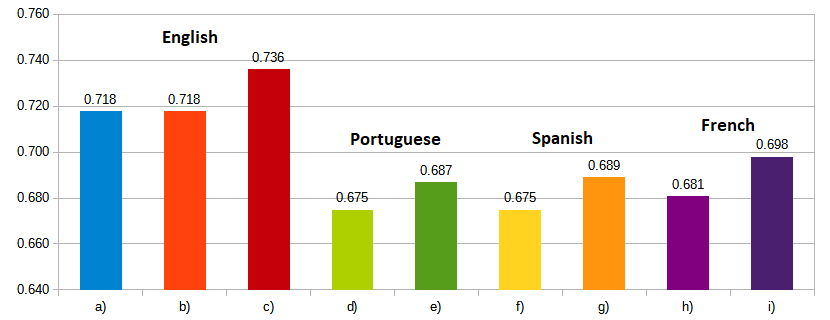}
\caption{Exact Set Match without Values for multilingual inferences:\newline
a) BART-large trained in English dataset standard, inferred in English (baseline);\newline
b) mT5-large model trained in English, Portuguese, Spanish and French standard quad dataset, inferred in English;\newline
c) mT5-large model trained in English, Portuguese, Spanish and French \textbf{FIT quad dataset}, inferred in English;\newline
d) mT5-large model trained in English, Portuguese, Spanish and French standard quad dataset, inferred in Portuguese;\newline 
e) mT5-large model trained in English, Portuguese, Spanish and French \textbf{FIT quad dataset}, inferred in Portuguese;\newline
f) mT5-large model trained in English, Portuguese, Spanish and French standard quad dataset, inferred in Spanish;\newline 
g) mT5-large model trained in English, Portuguese, Spanish and French \textbf{FIT quad dataset}, inferred in Spanish;\newline
h) mT5-large model trained in English, Portuguese, Spanish and French standard quad dataset, inferred in French;\newline 
i) mT5-large model trained in English, Portuguese, Spanish and French \textbf{FIT quad dataset}, inferred in French..} \label{fig3}
\end{figure}

The mT5-large fine-tuned with the quad dataset can infer questions in each of the four languages trained. Figure~\ref{fig3} shows the results for the exact set match without values, for inferences with the validation dataset in English (Figure~\ref{fig3}b and \ref{fig3}c), translated into Portuguese (Figure~\ref{fig3}d and \ref{fig3}e), Spanish (Figure~\ref{fig3}f and \ref{fig3}g) and French (Figure~\ref{fig3}h and \ref{fig3}i). These results were produced with mT5-large fine-tuned with the standard quad (English, Portuguese, Spanish and French) Spider train dataset Figure~\ref{fig3}b, \ref{fig3}d, \ref{fig3}f and \ref{fig3}h. The mT5-large fine-tuned with FIT quad (English, Portuguese, Spanish and French) Spider train dataset Figure~\ref{fig3}c, \ref{fig3}e, \ref{fig3}g and \ref{fig3}i. For all the results, the checkpoint that produced the best result for each language was selected.

Languages different from English have lower results, which can be attributed to the pre-training of the model mT5~\cite{mT52021}. It used much more tokens in English (2,733 B) than in Portuguese(146 B), Spanish(433 B), and French(318 B). Another aspect, English words in the question represent tables and columns names with the same words;  for other languages, the words in the question represent tables and columns names in English, once we preserved the English language in the Database Schema.
The mT5 model fine-tuned with the FIT quad training dataset shows better inference for all 4 languages if compared to the mT5 model fine-tuned with the standard FIT quad training dataset. This reinforces the importance of schema pruning.


\section{Conclusion and Future Work}

This work introduced new ideas to NL2SQL, particularly for multilingual settings.
For exact set match accuracy, proposed techniques increased standard metrics from 0.718 to 0.736 with the Dev validation dataset. Note that the original Spider is not used entirely,  because mRAT-SQL, which borrows the same code from RAT-SQL+GAP, drops examples that exceed 512 tokens. 
To prove the hypotheses that pruning the schema tables and columns names will help the training processes with more examples, we manually performed this pruning and created a FIT version of the Spider dataset that does not have any examples excluded, this allowed the self-attention transformer mechanism to treat the entire training dataset. This Spider FIT dataset version can easily plug in other techniques that use the Spider dataset.
The next step is to plug the Spider FIT dataset in another technique to evaluate the results.

\section*{Abbreviations}

DEV\quad Validation dataset\\
ETC\quad Extended transformer construction\\
GAP\quad Generation-augmented pre-training\\
mT5\quad Multilingual text-to-text transfer transformer model\\
NLP\quad Natural language processing\\
NL2SQL\quad Natural language to SQL\\
LGESQL\quad Line graph enhanced text-to-SQL\\
PICARD\quad Parsing incrementally for constrained auto-regressive decoding\\
RAT-SQL\quad Relation-aware transformer SQL\\
SQL\quad Structured query language\\
S²SQL\quad Syntax to question-schema graph encoder for text-to-SQL\\
T5\quad Text-to-text transfer transformer model\\

\section*{Declarations}

\begin{itemize}
\item \textbf{Funding} This work was carried out at the Center for Artificial Intelligence (C4AI-USP), with support by the Sao Paulo Research Foundation (FAPESP grant \#2019/07665-4) and by the IBM Corporation. The second author is partially supported by Conselho Nacional de Desenvolvimento Cientifico e Tecnologico (CNPq), grant 312180/2018-7.
\item \textbf{Conflict of interest/Competing interests} The authors have no relevant financial or non-financial interests to disclose.
\item \textbf{Ethics approval and consent to participate} Not applicable.
\item \textbf{Consent for publication} Not applicable.
\item \textbf{Availability of data and materials} \underline{https://github.com/C4AI/gap-text2sql}
\item \textbf{Code availability} \underline{https://github.com/C4AI/gap-text2sql} 
\item \textbf{Authors' contributions} The two authors had an equivalent contribution to the paper write. 
\end{itemize}

\end{document}